\documentclass[letterpaper, 10 pt, conference]{ieeeconf}  

\IEEEoverridecommandlockouts                              

\usepackage{fullpage,graphicx,psfrag,amsmath,amsfonts,verbatim,tikz}

\usepackage[small,bf]{caption}
\usepackage{xcolor}
\usepackage{hyperref}
\usepackage{tikz}
\usepackage{listings}
\usepackage{bbm}
\usepackage{fancyvrb}

\newcommand{\inR}{\in \mathbb{R}}
\newcommand{\CFree}{\mathcal{C}^\mathrm{free}}
\newcommand{\Cfree}{\CFree}

\newcommand{\minz}{\mathop{\text{minimize} }}
\newcommand{\maxz}{\mathop{\text{maximize} }}

\newcommand{\calB}{\ensuremath{\mathcal{B}}}

\newcommand{\calE}{\ensuremath{\mathcal{E}}}

\newcommand{\calG}{\ensuremath{\mathcal{G}}}
\newcommand{\calH}{\ensuremath{\mathcal{H}}}

\newcommand{\calK}{\ensuremath{\mathcal{K}}}

\newcommand{\calR}{\ensuremath{\mathcal{R}}}

\newcommand{\calT}{\ensuremath{\mathcal{T}}}

\newcommand{\calV}{\ensuremath{\mathcal{V}}}

\newcommand{\setR}{\ensuremath{\mathbb{R}}}

\newcommand{\subjectto}{\mathop{\text{subject to}}}

\newcommand{\Vol}{\mathop{\bf vol}}

\usepackage[utf8]{inputenc}

\DeclareFixedFont{\ttb}{T1}{txtt}{bx}{n}{10} 
\DeclareFixedFont{\ttm}{T1}{txtt}{m}{n}{10}  

\usepackage{color}
\definecolor{deepblue}{rgb}{0,0,0.5}
\definecolor{deepred}{rgb}{0.6,0,0}
\definecolor{deepgreen}{rgb}{0,0.5,0}

\usepackage{listings}

\definecolor{codegreen}{rgb}{0,0.6,0}
\definecolor{codegray}{rgb}{0.5,0.5,0.5}
\definecolor{codepurple}{rgb}{0.58,0,0.82}
\definecolor{backcolour}{rgb}{0.95,0.95,0.92}

\lstdefinestyle{mystyle}{
	backgroundcolor=\color{backcolour},   commentstyle=\color{codegreen},
	keywordstyle=\color{magenta},
	numberstyle=\tiny\color{codegray},
	stringstyle=\color{codepurple},
	basicstyle=\ttfamily\footnotesize,
	breakatwhitespace=false,         
	breaklines=true,                 
	captionpos=b,                    
	keepspaces=true,                 
	numbers=left,                    
	numbersep=5pt,                  
	showspaces=false,                
	showstringspaces=false,
	showtabs=false,                  
	tabsize=2
}
\newcommand{\ps}{{\sc{\textbf{Min $\alpha$-ApproxConvexCover}}}~}
\newcommand{\maxclique}{\textsc{\textbf{MaxClique}}\xspace}
\newcommand{\mcc}{\textsc{\textbf{MinCliqueCover}}\xspace}
\newcommand{\marcsthingy}{\text{IOS}}

\usepackage[most]{tcolorbox}
\usepackage[ruled,vlined]{algorithm2e}
\usepackage{algorithmicx}
\usepackage{amsmath, amsfonts, bm, amssymb, tabularx, mathtools, amsthm}
\usepackage{afterpage}

\newtheoremstyle{mytheorem}
  {\topsep}
  {\topsep}
  {\itshape}
  {}
  {\bfseries}
  {}
  {.5em}
  {}
\theoremstyle{mytheorem}
\newtheorem{definition}{Definition}

\newenvironment{problem}[1]{%
  \begin{tcolorbox}[colback=white,colframe=gray!40,coltitle=black,title={\textbf{Problem: } \sc{\textbf{#1}}}]
}{%
  \end{tcolorbox}
}

\usepackage{siunitx}
\usepackage{multirow}
\usepackage{booktabs}
\usepackage{pifont}
\usepackage{caption}
\usepackage{subcaption}
\usepackage{verbatim}

\title{\LARGE \bf
Approximating Robot Configuration Spaces with few Convex Sets \\ using Clique Covers of Visibility Graphs
}

\author{
 Peter Werner, Alexandre Amice, Tobia Marcucci, Daniela Rus, and Russ Tedrake
\thanks{ All authors are with CSAIL, MIT. The corresponding author is P.Werner, \texttt{wernerpe@mit.edu}. This work was supported in part by Amazon.com PO\# \#2D-06310236, Air Force Research Laboratory FA8750-19-2-1000, Office of Naval Research (ONR), Awards No. N00014-22-1-2121 and N00014-18-1-2830, and the Toyota Research Institute (TRI). }%
}

\begin{document}
\bstctlcite{IEEEexample:BSTcontrol}

\maketitle

\thispagestyle{empty}
\pagestyle{empty}

\begin{abstract}
Many computations in robotics can be dramatically accelerated if the robot configuration space is described as a collection of simple sets. For example, recently developed motion planners rely on a convex decomposition of the free space to design collision-free trajectories using fast convex optimization. 
In this work, we present an efficient method for approximately covering complex configuration spaces with a small number of polytopes.
The approach constructs a visibility graph using sampling and generates a clique cover of this graph to find clusters of samples that have mutual line of sight.
These clusters are then inflated into large, full-dimensional, polytopes. We evaluate our method on a variety of robotic systems and show that it consistently covers larger portions of free configuration space, with fewer polytopes, and in a fraction of the time compared to previous methods.
\end{abstract}

\section{Introduction}
\begin{figure}[t]
\centering
\includegraphics[width=0.95\linewidth, trim={0.9cm 0cm 2.0cm 0cm}, clip]{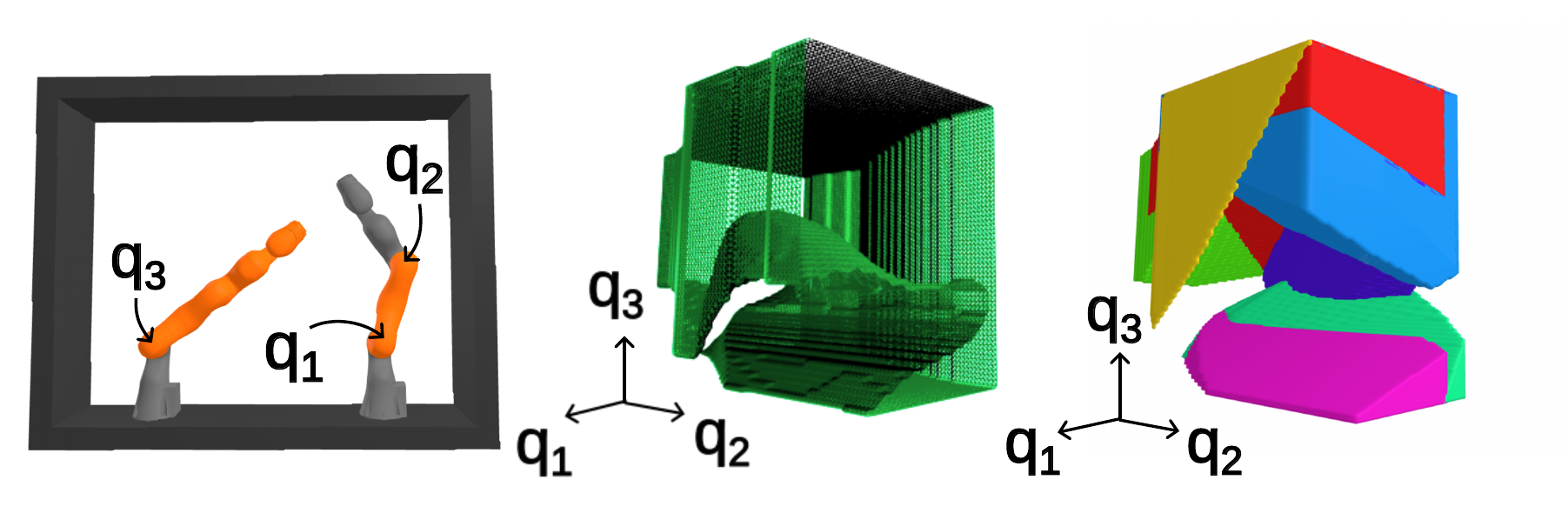}
\caption{The collision-free configuration space of a simple robot is decomposed into 7 polytopes, achieving around 92\% coverage. \textit{Left}: Robot with with 3 revolute joints $q_1$ to $q_3$.  \textit{Center}: Visualization of the full collision-free configuration space $\CFree$, given by the interior of the green mesh.  \textit{Right}: Approximate convex cover of $\CFree$ generated with the proposed method. See also: \href{https://sites.google.com/view/cspacevcc/home}{\textit{\underline{Website}}}, \href{https://www.youtube.com/watch?v=x37fPVST6Zk}{\textit{\underline{video}}}.
\vspace{-0.5cm}}
\label{fig:titlefigure}
\end{figure}

Approximating complex sets as a union of simpler sets is a common pre-processing method for accelerating downstream computations.
Familiar examples include clustering methods for high-dimensional data in machine learning \cite[\S2.5.2]{bishop2006pattern}, approximating complex shapes via triangular meshes to facilitate graphics rendering \cite{pharr1997rendering}, and describing geometries as unions of convex sets for efficient collision checking \cite{gilbert1988fast}.
Similarly, recently developed methods for robot motion planning rely on (conservative) decompositions of the environment into convex sets to design smooth trajectories around obstacles using efficient convex optimization~\cite{marcucci2021shortest,marcucci2022motion,cohn2023non,kurtz2023temporal,marcucci2023fast}.
These motion planners have demonstrated great potential, and performance frequently superior to widely used sampling-based methods~\cite{marcucci2022motion}.
However, the required decomposition of the environment is a daunting task, often demanding a substantial degree of human supervision. This is largely due to the fact that the collision-free subset of a robot's configuration space ($\CFree$) is intractable to describe analytically~\cite[\S3]{latombe2012robot}, even if the robot's task space is relatively simple (e.g. see Figure~\ref{fig:titlefigure}).

In~\cite{deits2015computing}, the authors proposed an algorithm called IRIS to quickly compute large convex subsets of $\CFree$.
This method takes as input a ``seed'' configuration and inflates a large polytopic region around this seed point using convex optimization. While originally limited to the case of convex obstacles, the IRIS algorithm has been recently extended to handle nonconvex obstacles and the complicated configuration spaces of multi-link robot manipulators~\cite{amice2022finding,petersen2023growing,dai2023certified}.
The polytopes generated with these algorithms have been used with great success for motion planning in high dimensions~\cite{marcucci2022motion,cohn2023non,kurtz2023temporal}.
Nevertheless, seeding collections of these regions so that they cover diverse areas of $\CFree$ remains a challenge: manual seeding is tedious, and naively growing regions around randomly chosen configurations leads to very inefficient decompositions.

 \begin{figure*}[t]
\centering
\includegraphics[width = 0.95\textwidth]{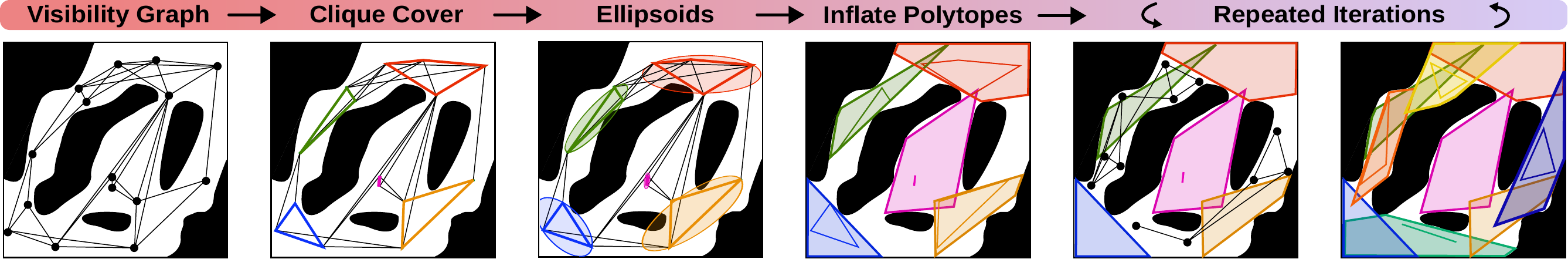}
\caption{Sketch of the proposed algorithm on a simple example. \textit{First four figures}: Samples are drawn uniformly from $\CFree$ to build a visibility graph. The visibility graph is decomposed into five cliques. The principal directions and locations of the cliques are used to direct a region-inflation algorithm. \textit{Remaining two figures}: This process is repeated until sufficient coverage is obtained by drawing new samples from the remaining free space, and repeating the previous steps.
\vspace{-0.4cm}
}
\label{fig:algocomic}
\end{figure*}

In this paper, we propose an efficient method for the approximate decomposition of robot configuration spaces into few large convex sets, without any human supervision. A guiding illustration is shown in Figure \ref{fig:algocomic}.

Similar to some motion-planning algorithms~\cite{simeon2000visibility,lozano1979algorithm}, our method constructs a visibility graph by sampling points in $\CFree$.
The vertices of this graph are collision-free samples, and the edges connect pairs of points with mutual line of sight.
The visibility graph contains rich information about the geometry of $\CFree$.
In particular, our key observation is that, as the number of samples grows, fully connected subgraphs of this visibility graph (so-called ``cliques'') tend to represent better and better approximations of collision-free convex sets in the underlying configuration space.
Our approach is to decompose the visibility graph into a small collection of large cliques.
We then circumscribe the points in each clique with an ellipsoid by solving a convex-optimization problem.
The center and the principal directions of these ellipsoids are subsequently used to initialize an inflation algorithm analogous to IRIS.

Through a large variety of experiments, we show that our algorithm outperforms previous approaches for seeding and inflating convex regions; both in terms of runtimes and number of regions used to cover the space.
As an example, for a robot arm with seven degrees of freedom, and many task-space obstacles, our method requires approximately $46$ regions and one hour of computations to cover $70\%$ of $\CFree$, whereas the approach outlined in~\cite{petersen2023growing} requires ten times more regions and is ten times slower.



\section{Related Works}\label{sec:relworks}
Finding the minimum convex cover of a set is a hard problem; even in the case of two-dimensional polygons, the problem is hard to solve exactly and approximately.\footnote{More formally, the problem is $\exists\mathbb{R}$-complete \cite{abrahamsen2022covering}, and therefore NP-hard, as well as APX-hard~\cite{eidenbenz2003approximation}.}
Nonetheless, finding low-cardinality convex covers of high-dimensional nonconvex spaces (both polygonal and non-polygonal) remains a problem of practical importance, for which a variety of approximate algorithms have been devised.
In the following, we group these algorithms into two categories: ones that require explicit (e.g., analytic) descriptions of $\CFree$, and ones that only use implicit descriptions of $\CFree$ (and are hence suitable for most complex configuration spaces). In this literature review, we particularly focus on IRIS algorithms, due to their efficient scaling to high dimensions. 

\subsection{Algorithms Requiring Explicit Obstacle Descriptions}\label{ssec:algo_explicit}
The recent work~\cite{abrahamsen2023constructing} constructs low-cardinality convex covers of two-dimensional polygons by first decomposing them into small convex pieces (for example triangles), and then by joining the pieces based on a small clique cover of an approximated set-to-set visibility graph. Similarly, the visibility graph of polygons is directly used to construct convex covers in \cite{da2023shadoks}.
In three dimensions the approach in~\cite{mamou2009simple} can be used to decompose $\CFree$ into sets that are approximately convex, provided that a triangle mesh description is available.
Finally, if the configuration-space obstacles are explicitly described as convex sets, the original IRIS algorithm from~\cite{deits2015computing} can be used to inflate a large polytope in $\Cfree$ around a specified seed point in arbitrary dimensions.

\subsection{Algorithms Allowing Implicit Obstacle Descriptions}\label{ssec:algo_implicit}
Most commonly, the explicit descriptions of the obstacles are only available in task space; while the collision-free configuration space $\CFree$ is defined implicitly through the robot's inverse kinematics, and is intractable to describe analytically~\cite[\S3]{latombe2012robot}.
 Some works have focused on approximately describing the configuration space obstacles, rather than $\CFree$~\cite{varava2021free, kavraki1995computation, branicky1990computing}. 
Conversely, multiple works have studied directly obtaining convex decompositions of $\CFree$.
In~\cite{sarmientoy2005sample} visibility graphs and kernels are used to compute convex decompositions of three-dimensional spaces via sample-based collision checking.
However, their method represents the convex sets through their vertices, and is inefficient in higher dimensions.
In the family of IRIS algorithms, two methods can deal with implicit descriptions of $\CFree$: IRIS-NP~\cite{petersen2023growing} and C-IRIS~\cite{amice2022finding,dai2023certified}.
The former extends the original IRIS method~\cite{deits2015computing} to arbitrary configuration spaces using nonlinear programming and inflates polytopes that are collision-free with high probability.
The latter grows polytopes that are rigorously \emph{certified} to be collision-free using a rational reparametrization of the configuration space and sums-of-squares programming. 

\section{Convex Covers, Visibility, and Cliques}\label{sec:definitions}

In this section, we formally define our main problem: approximating the free configuration space $\CFree$ with a low-cardinality collection of convex sets. We also briefly review the main technical tools that we will use in the development of our algorithm, namely, visibility graphs and clique covers.

\subsection{Problem Statement}
Let $\CFree \subseteq \setR^{n}$ be the collision-free subset of an $n$-dimensional configuration space, which we assume to have a well-defined finite volume.
Let also $\alpha$ be a constant in the interval $(0,1]$. 

\begin{definition}
An \emph{$\alpha$-approximate convex cover} of $\CFree$ is a collection of potentially overlapping convex sets $\calR_1, \ldots, \calR_N \subseteq \CFree$ whose union covers at least an $\alpha$-fraction of the volume of $\CFree$:
$$
\Vol\left(\bigcup_{i=1}^{N} \calR_{i}\right) \geq \alpha\Vol\left(\CFree\right).
$$
\vspace{-0.2cm}
\end{definition}

Our problem is to find an $\alpha$-approximate convex cover of minimum cardinality $N$.

\begin{problem}{\ps}
  \vspace{-0.2cm}
  \begin{gather*}
    \minz N \subjectto \\
    \Vol\left(\bigcup_{i=1}^{N} \calR_{i}\right) \geq \alpha\Vol\left(\CFree\right), \\
    \calR_{i} \subseteq \CFree, \quad \forall i = 1, \dots, N.
  \end{gather*}
  \vspace{-0.6cm}
\end{problem}

In practice,  we are interested in solving this problem for values of $\alpha$ that are sufficiently high to accomplish a task of interest, such as collision-free motion planning, but not so large that the cardinality $N$ grows unreasonably. Indeed, when $\alpha = 1$, \ps might not even have a finite solution.\footnote{For $\alpha<1$ a finite solution is guaranteed to exist. This can be seen by computing the volume of $\CFree$ with a lower Darboux integral over finite hyperrectangle partitions. }

\subsection{Visibility Graphs and Clique Covers}
Our algorithm is based on the idea that clusters of points that see each other can approximate convex subsets of $\CFree$. Here, we formally define the notion of visibility as well as introduce some formal tools from graph theory to guide the development of our algorithm.

We begin by defining visibility in $\CFree$.
\begin{definition}\label{D: visibility}
Two points $q,q' \in \CFree$ are said to  \emph{see each other} if the entire line connecting them is collision-free: $tq + (1-t)q' \in \CFree$ for all $t \in [0,1]$.
Notice that this definition is symmetric in $q$ and $q'$.
\end{definition}

We are now ready to define the visibility graph of a set of collision-free configurations.
\begin{definition}\label{D: visibility graph}
The \emph{visibility graph} of a set of points $q_1, \dots, q_K \in \CFree$ is an undirected graph $\calG = (\calV, \calE)$ with vertices $\calV = \{1, \ldots, K\}$, and with an edge $\{i,j\} \in \calE$  for every pair of distinct points $q_i$ and $q_j$ that see each other. 
\end{definition}

We show an example of a visibility graph in Figure~\ref{fig:algocomic}. We note that clusters of points that can all see each other form a clique in the visibility graph.

\begin{definition}
  Let $\calG = (\calV, \calE)$ be an undirected graph. A \emph{clique} $\calK$ is a subset of $\calV$ where every pair of vertices is connected by an edge. 
\end{definition}

Note that if a cluster of configurations can be placed in the same convex set, then these configurations must form a clique in the visibility graph. The second panel in Figure~\ref{fig:algocomic} highlights a collection of five cliques in the visibility graph that have this property. These five cliques form what is called a clique cover; which resembles a discrete analog of a convex cover.

\begin{definition}
  A collection $\calT$ of cliques $\calK_{1}, \dots, \calK_{N}$ is a \emph{clique cover} of a graph $\calG$ if every vertex in the graph is contained in at least one clique.
\end{definition}  

A natural discrete counterpart of the minimum convex cover is the \mcc problem, where we look for the minimum number of cliques $N$ required to cover a graph. Our observation is that, as the number of samples in the visibility graph increases, a minimum clique cover typically does an increasingly better job of approximating a minimum convex cover.
Limitations of this analogy are discussed in \S\ref{sec:discussion}.

\mcc is NP-complete~\cite{karp2010reducibility}.
There exist heuristics, such as \cite{strash2022effective}, that attempt to solve \mcc directly.
Alternatively, one can greedily construct a clique cover by repeatedly eliminating the largest clique. The problem of finding the largest clique in a graph is called \maxclique. Even though this problem is also NP-complete~\cite{karp2010reducibility}, it is often substantially faster to solve in practice. 
We found the latter approach with exact solutions of \maxclique to perform particularly well on our problem instances.





\section{Algorithm}\label{sec:algorithm}

We now present our Visibility Clique Cover (VCC) algorithm, which consists of four main steps.
First, we randomly sample a collection of points in $\CFree$ and construct their visibility graph.
Second, we compute an approximate clique cover of the graph.
Third, we summarize the geometric information of each clique using an ellipsoid.
Fourth, we use these ellipsoids to initialize a polytope-inflation algorithm analogous to IRIS.
This process is repeated until the generated set of polytopes $\calR$  covers a given fraction $\alpha$ of $\CFree$.
This procedure is summarized in Algorithm~\ref{alg:cseeding}, and the remainder of this section details the individual steps.

\vspace{-0.2cm}
\begin{algorithm}
\SetAlgoLined
\caption{\textsc{VisibilityCliqueCover}}
 \label{alg:cseeding}
\SetKwInOut{Input}{Input}
\SetKwInOut{Output}{Output}
\SetKw{KWAlgorithm}{Algorithm:} 
\Input{

$\alpha$: coverage threshold

$K$: number of samples per iteration

$s_\text{min}$: minimum clique size
}
\Output{

$\calR$: set of convex polytopes approximating $\CFree$
}

\KWAlgorithm{}

$\calR \gets \emptyset$

\While{$\textsc{CheckCoverage}(\calR)\leq \alpha$}{
$\calG ~\gets \textsc{SampleVisibilityGraph}(K, \calR)$ \\
$\calT \gets \textsc{TruncatedCliqueCover}(\calG, s_\text{min})$\\
$\calB \gets \textsc{MinVolumeEllipsoids}(\calT)$\\
$\calR \gets \calR \cup \textsc{InflatePolytopes}(\calB)$
}

\Return $\calR$
\end{algorithm}
\vspace{-0.3cm}
\subsection{Sampling the Visibility Graph}

At the beginning of every iteration of VCC, the subroutine $\textsc{SampleVisibilityGraph}$ 
samples $K$ configurations uniformly at random from the portion of $\CFree$ that is not already covered by the polytopes in $\calR$.
Then, it constructs the visibility graph $\calG = (\calV, \calE)$, by checking for collisions along the line segments connecting each pair of sampled configurations.
Currently, this is performed using sampling-based collision checkers.
Exact visibility checking is possible using methods such as \cite{schwarzer2004exact},~\cite[\S5.3.4]{lavalle2006planning}, or~\cite{amice2023certifying}.

\subsection{Truncated Clique Cover}\label{ssec:trunccliquecov}

In the subroutine \textsc{TruncatedCliqueCover} we approximately cover the visibility graph with a collection of cliques, each of which contain at least $s_{\min}$ vertices.
We construct this approximate cover $\calT$ greedily, by solving a sequence of \maxclique problems.
Each instance of \maxclique is formulated as an integer linear program
\begin{subequations}
\label{eqn:maxclique}
\begin{align}
\maxz \quad & \sum_{i = 1}^K b_i\\
\subjectto \quad & b_i + b_j \leq 1, && \forall \{i,j\} \in \bar \calE, \\
& b_i \in \{0,1\}, && \forall i=1,\ldots, K.
\end{align}
\end{subequations}
A binary decision variable $b_i$ is added for each vertex. The role of this variable is to take unit value if and only if vertex $i$ is included in the clique.
The set $\bar \calE$ contains all the pairs of vertices $\{i,j\}$ such that $i \neq j$ and $\{i,j\} \notin \calE$.
Therefore the first constraint ensures that two vertices are selected only if they share an edge.

After solving the integer program~\eqref{eqn:maxclique}, the clique found is removed from the graph and added to the clique cover.
Since small cliques are not informative, we stop this iterative process when the largest clique left in the graph is smaller than a given threshold
$s_{\min}$.
For this reason, our clique covers will be truncated, i.e., generally, not all vertices will be contained in one of the cliques.

\subsection{Summarizing Cliques with Ellipsoids}\label{ssec: ellipsoids}

In the subroutine \textsc{MinVolumeEllipsoids}, we solve a semidefinite program to enclose each clique with an ellipsoid of minimum volume~\cite[\S8.4.1]{boyd2004convex}.
This collection $\calB$ of ellipsoids allows us to summarize the geometry of each clique with a point and a set of principal directions, which are then used to initialize the region-inflation algorithm.
For the upcoming computations, it is necessary that the center of each ellipsoid is not in collision; if this is not the case, we recenter the ellipsoid around the vertex in the clique that is closest to its center.

 \subsection{Inflating Polytopes}\label{ssec:IRIS}

\begin{figure}
\centering
\includegraphics[width = 0.28\linewidth,trim = {3.5cm 3.5cm 3.4cm 3.5cm}, clip]{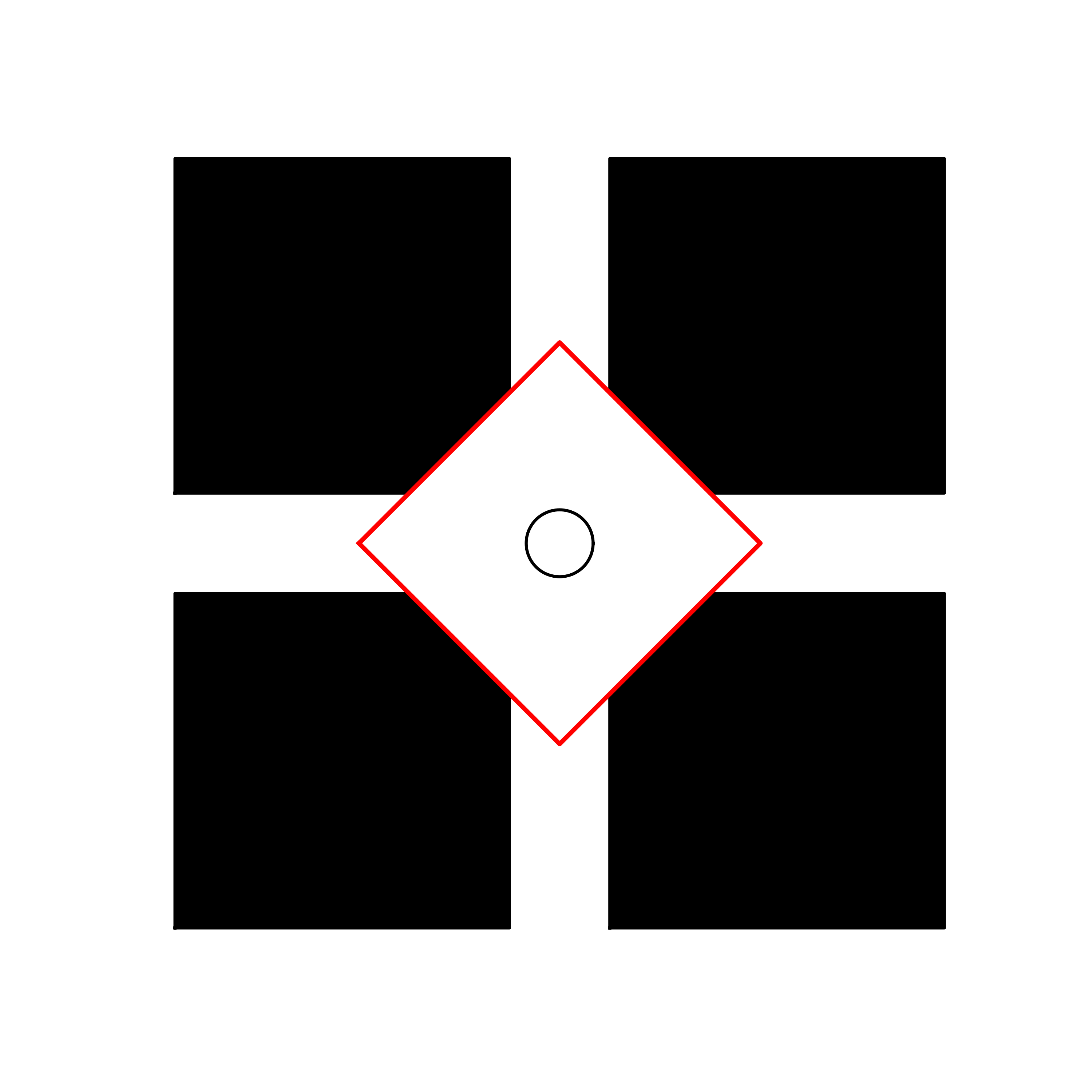}
\hfill
\includegraphics[width = 0.28\linewidth,trim = {3.5cm 3.5cm 3.4cm 3.5cm}, clip]{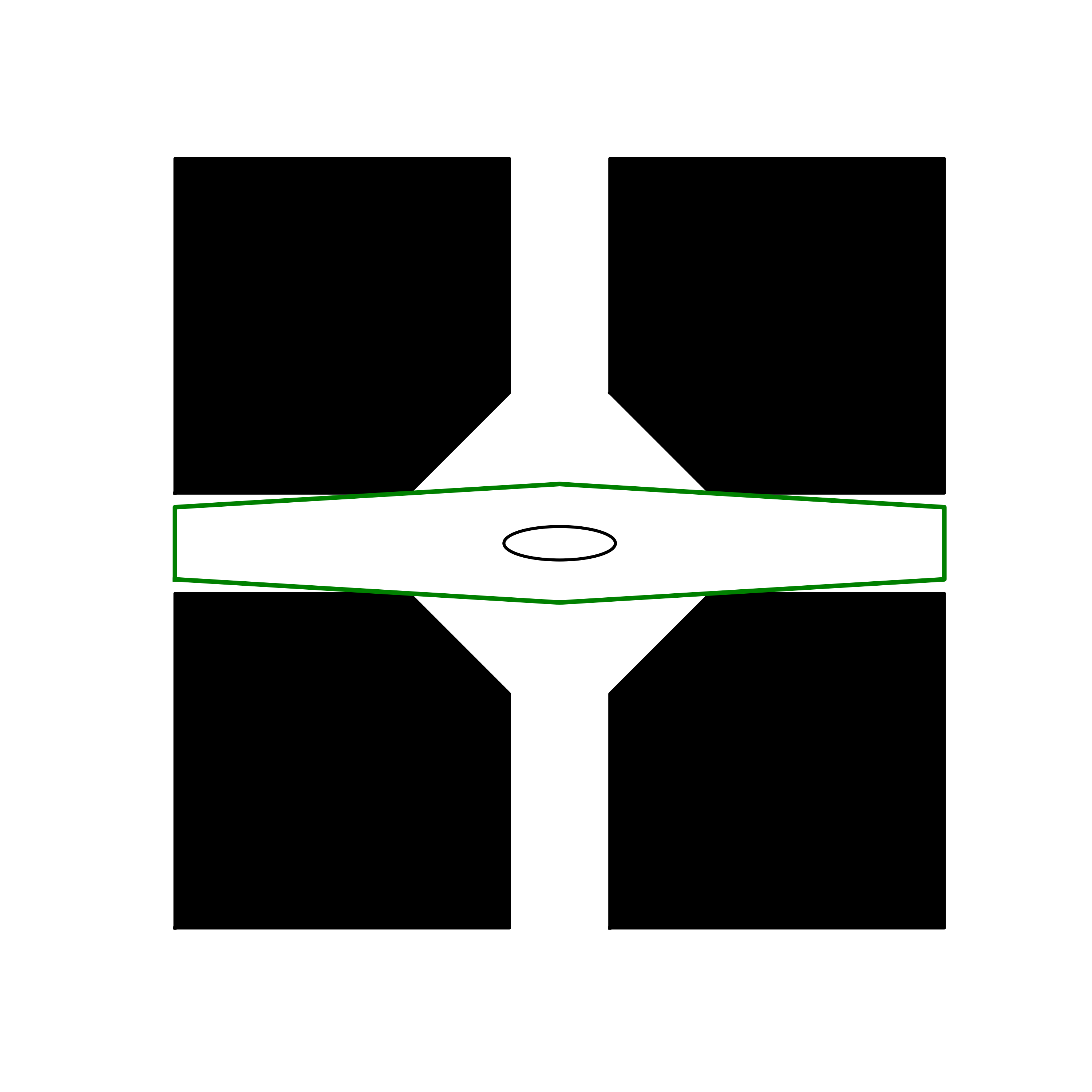}
\hfill
\includegraphics[width = 0.28\linewidth,trim = {3.5cm 3.5cm 3.4cm 3.5cm}, clip]{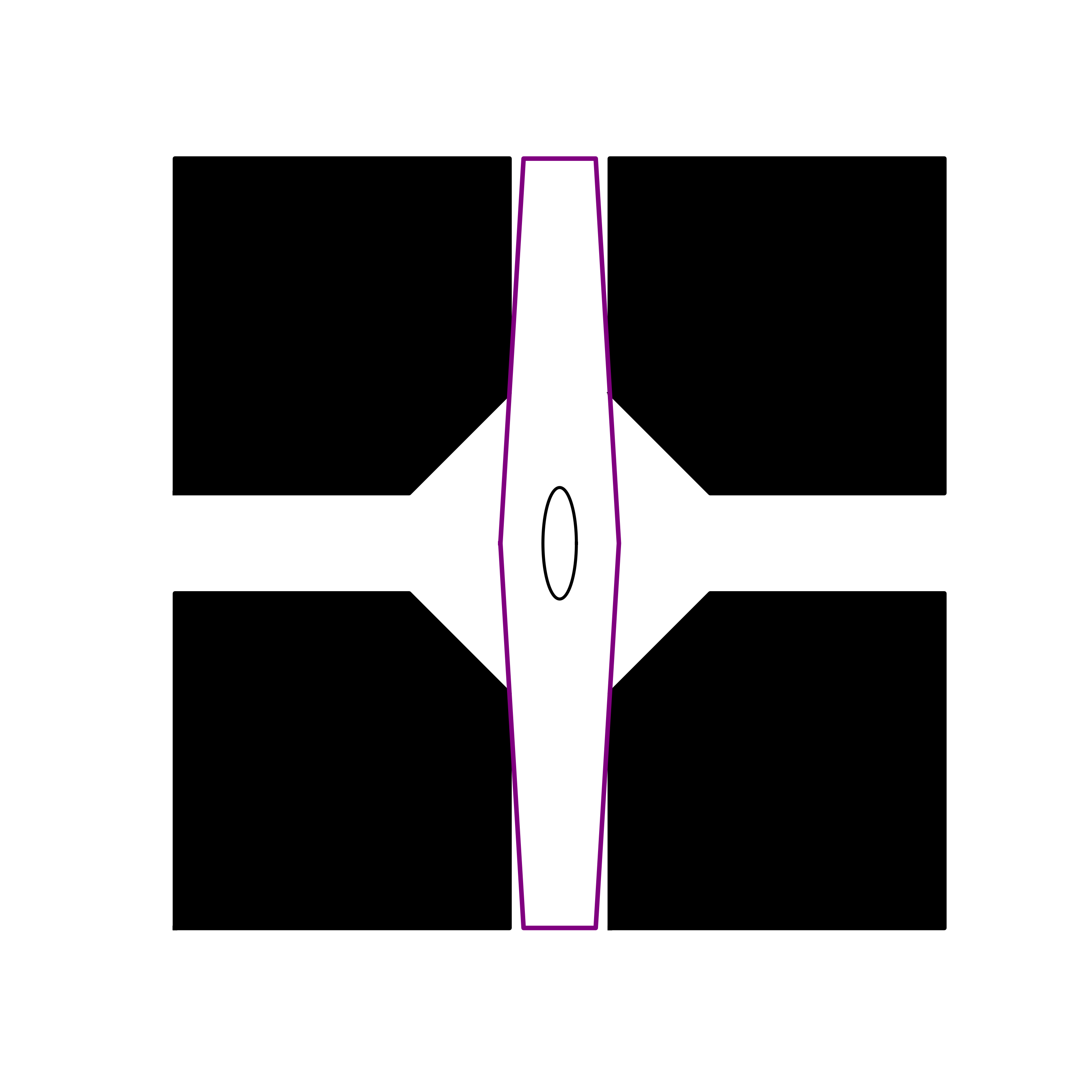}
\caption{The growth direction of an IRIS region can be guided by the initial distance metric.
IRIS is initialized with three ellipsoids with same center but different principal axes, resulting in polytopes that cover different portions of $\Cfree$.}
\label{fig:directional iris growth}
\vspace{-0.5cm}
\end{figure}

In the last step of VCC, the subroutine \textsc{InflatePolytopes} inflates a large, collision-free polytope around the center of each ellipsoid induced by a clique.

Let us initially assume that the obstacles are convex.
Consider a single ellipsoid;
using convex optimization, we compute the point in each obstacle that is closest to the center of the ellipsoid, according to the distance metric induced by the ellipsoid.
These points anchor separating hyperplanes between the ellipsoid center and the obstacles, which form a polytope of obstacle-free space. These steps are repeated for each ellipsoid (i.e., for each clique) to obtain a collection of collision-free polytopes that we add to the set $\calR$.
\begin{figure*}
    \centering
    \begin{subfigure}{0.23\textwidth}
        ~~\includegraphics[width = 0.8\linewidth,trim = {0cm 1cm 1cm 1cm}, clip]{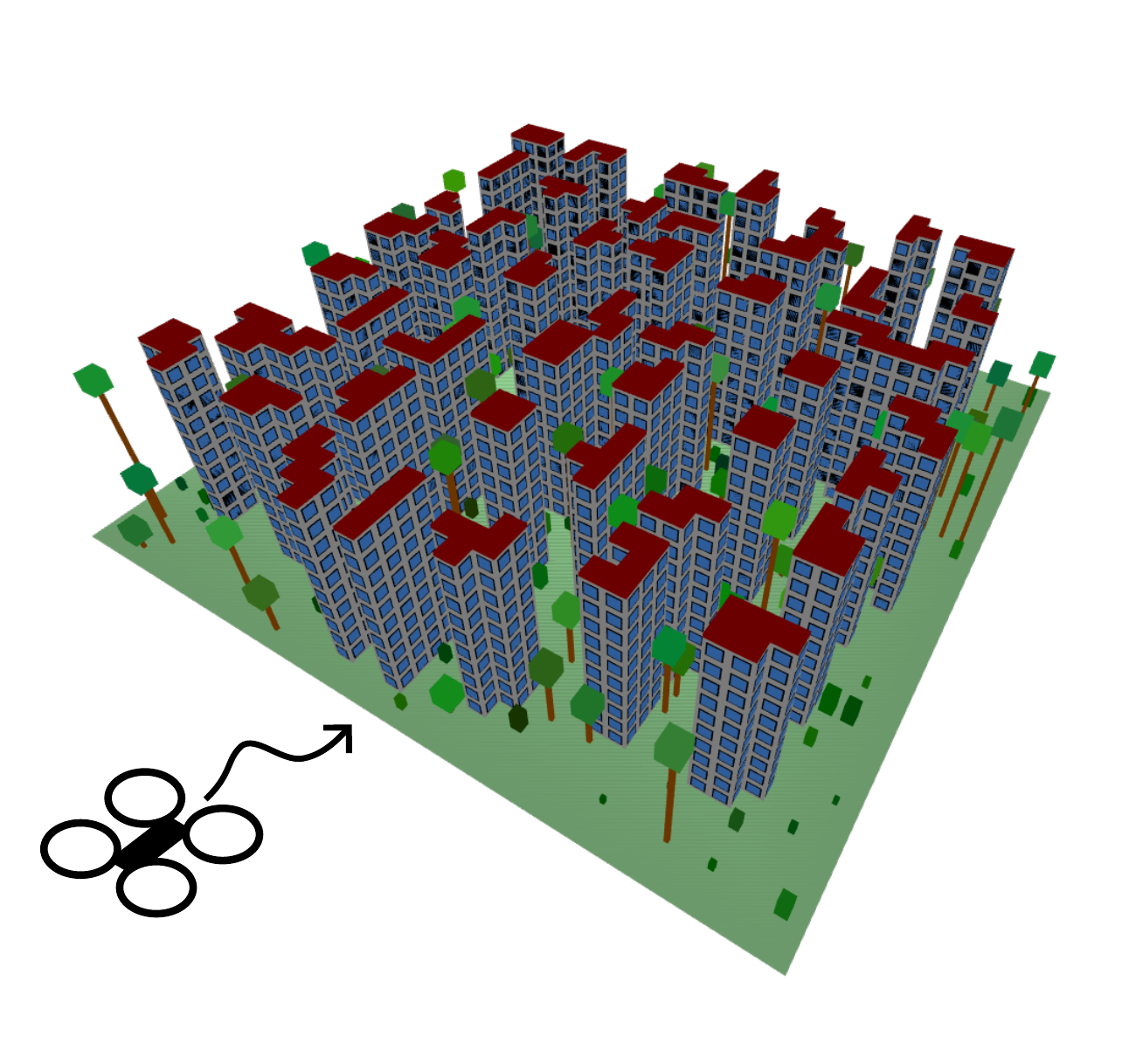}
        \caption{\href{https://wernerpe.github.io/files/Village.html}{\underline{\texttt{Village}}}}
    \end{subfigure} 
    \begin{subfigure}{0.23\textwidth}
    \centering
        \includegraphics[width = 0.75\linewidth,trim = {0.8cm 0.8cm 0.8cm 1cm}, clip]{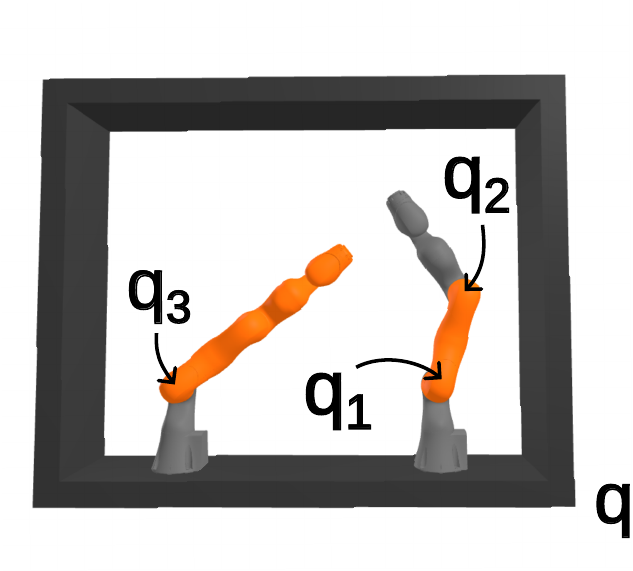}
    \caption{\href{https://wernerpe.github.io/files/3DOFFLIPPER_arxiv.html}{\underline{\texttt{3DOF Flipper}}}}
    \end{subfigure}
    \begin{subfigure}{0.23\textwidth}
    \centering
        \includegraphics[width = 0.75\linewidth]{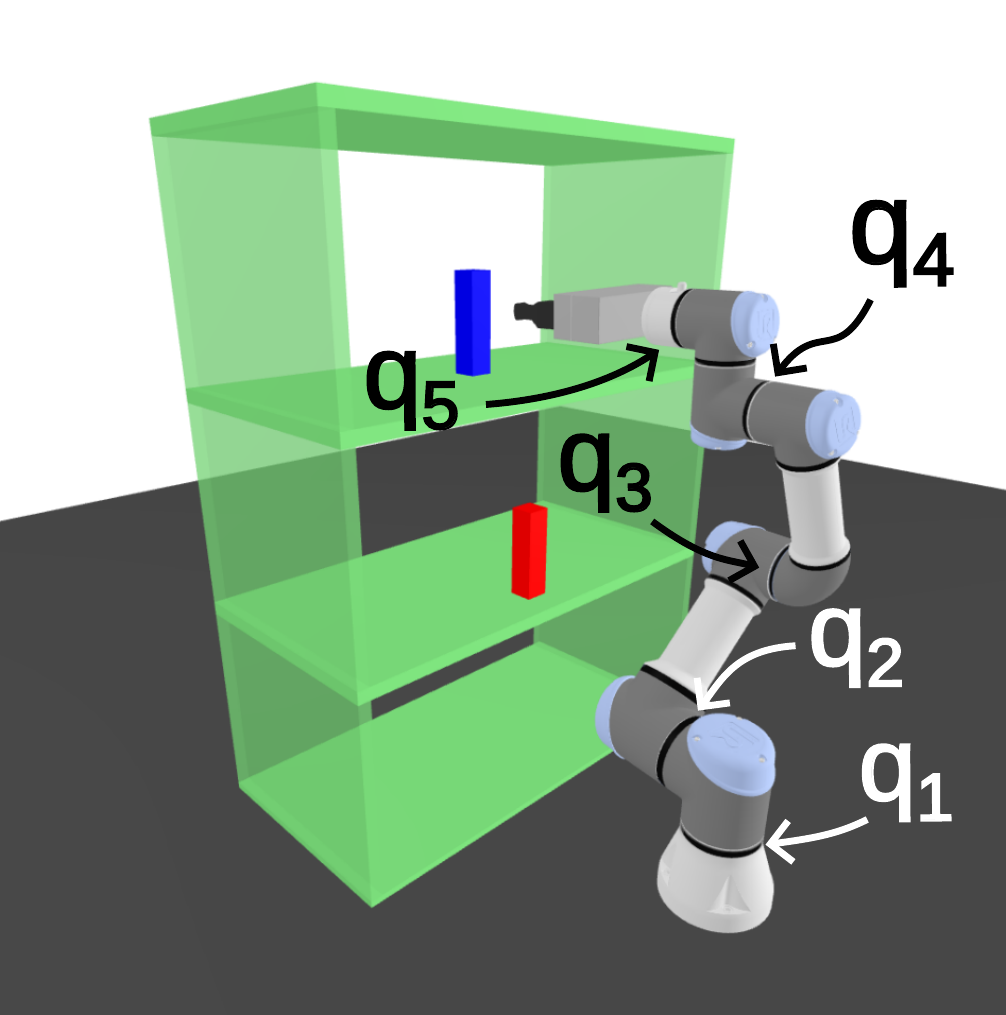}
        \caption{\href{https://wernerpe.github.io/files/5DOFUR5_arxiv.html}{\underline{\texttt{5DOF UR3}}}}
    \end{subfigure} 
    \begin{subfigure}{0.23\textwidth}
    \centering
        \includegraphics[width = 0.75\linewidth]{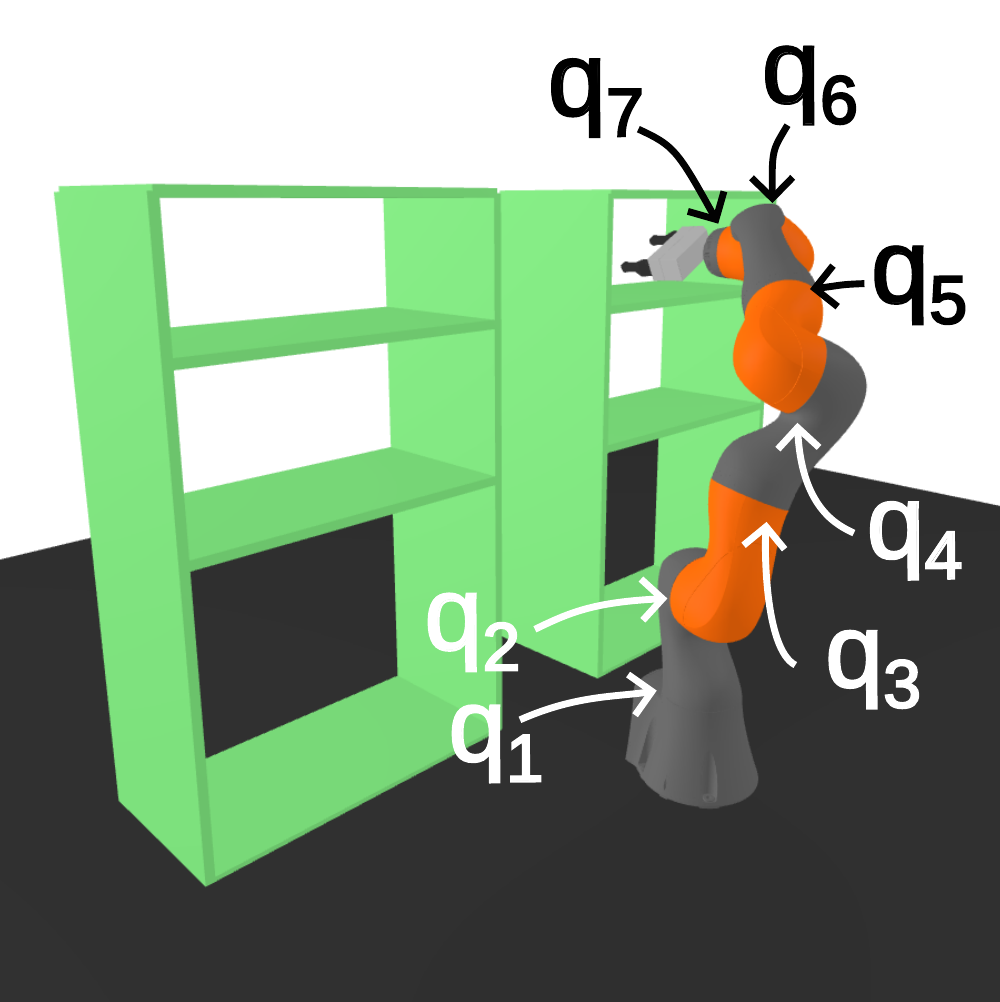}
        \caption{\href{https://wernerpe.github.io/files/7DOF_IIWA_arxiv.html}{\underline{\texttt{7DOF IIWA}}}}
    \end{subfigure}
    \vspace{-0.3cm}
    \end{figure*}
\begin{table*}[]
    \centering
    \resizebox{0.99\textwidth}{!}{
  \begin{tabular}{lcccccccc}
    \toprule
    Domain&
      \multicolumn{2}{c}{\texttt{Village}} &
      \multicolumn{2}{c}{\texttt{3DOF Flipper}} &
      \multicolumn{2}{c}{\texttt{5DOF UR3}} &
      \multicolumn{2}{c}{\texttt{7DOF IIWA}}\\
      Algorithm& {$\marcsthingy$} & {VCC (ours)} & {$\marcsthingy$} & {VCC (ours)} & {$\marcsthingy$} & {VCC (ours)} &{$\marcsthingy$} & {VCC (ours)}\\
      \midrule
      \# regions $|\calR|$ & 198.0\scriptsize{$\pm$13.6} & \textbf{93.9}\scriptsize{$\pm$7.5} & 10.4\scriptsize{$\pm$1.9}&\textbf{ 6.7}\scriptsize{$\pm$0.5} & 90.5\scriptsize{$\pm$18.8}&\textbf{35.1}\scriptsize{$\pm$1.6 }&482.6\scriptsize{$\pm$83.8}&\textbf{46.3}\scriptsize{$\pm$4.5} \\
    runtime [s] & 2.7e3\scriptsize{$\pm$2.9e2}& \textbf{1.7e3}\scriptsize{$\pm$3.4e2} & 2.0e2\scriptsize{$\pm$2.9e1}&\textbf{ 4.9e1}\scriptsize{$\pm$7.1 } & 1.12e4\scriptsize{$\pm$2.1e3}& \textbf{5.5e2}\scriptsize{$\pm$6.6e1}& 8.9e4\scriptsize{$\pm$1.9e4}& \textbf{5.7e3}\scriptsize{$\pm$  1.7e3} \\
    coverage threshold $\alpha$ & 0.8 & 0.8 & 0.9 & 0.9 & 0.75 & 0.75 & 0.7 &0.7 \\
    \# visibility vertices $K$ & - & 500 & - & 500 & - & 1000& - & 1500 \\
    min. clique size $s_\text{min}$ &-&10&-&10&-&10&-&20\\
    \bottomrule
  \end{tabular}
  }
\caption{Comparison of our Visibility Clique Cover (VCC) algorithm to the Iterative Obstacle Seeding ($\marcsthingy$) method from~\cite{petersen2023growing}, across four different environments. All experiments are repeated ten times. The numbers in the first two rows indicate the mean and the empirical standard deviation over the trials.
We observe that VCC achieves the given coverage targets with 1.6 to 10.4  times fewer regions, and between 1.4 to 20 times faster than IOS. 
   The environment names are linked to interactive visualizations.
   \vspace{-0.7cm}}
  \label{tab:results}
\end{table*}

These computations correspond to a single iteration of the IRIS algorithm~\cite{deits2015computing}, and ensure that the largest uniformly-scaled, collision-free version of the ellipsoid is contained in the resulting polytope. Figure~\ref{fig:directional iris growth} illustrates how the initial metric is fundamental in guiding the shape of the regions generated by IRIS.
Traditionally, the IRIS algorithm was initialized with an uninformed ellipsoidal metric and needed to run for multiple (expensive) iterations in order to expand and cover a larger volume of space; in VCC we require only a single IRIS iteration.



When the obstacles are not convex, we run one iteration of the nonlinear programming variant IRIS-NP~\cite{petersen2023growing} instead.
Alternatively, C-IRIS~\cite{amice2022finding,dai2023certified} could be employed to obtain certifiably collision-free regions.

\subsection{Convergence Check}

The subroutine \textsc{CheckCoverage} estimates the fraction of $\Cfree$ covered by the regions in $\calR$, and terminates our algorithm if this value exceeds the threshold $\alpha$.
Computing this fraction exactly is impractical, and so we resort to randomized methods.
The coverage is estimated by drawing a large number of $M$ samples in $\Cfree$, and computing the ratio of samples that land in at least one of the regions in $\calR$.
More sophisticated checks, such as one-sided Bernoulli hypothesis testing, are possible.

\subsection{Completeness}

Analogous to~\cite{kuffner2000rrt, kavraki1996probabilistic}, VCC is probabilistically complete under mild assumptions. This means as the number of iterations goes to infinity, the probability of completely covering $\CFree$ goes to one.
\section{Experiments}\label{sec:Experiments}
As there are no direct baseline methods, we compare our VCC algorithm against an extension of the method in~\cite[\S III.D]{petersen2023growing}. The natural extension of this approach is to iteratively grow polytopes around uniformly sampled points from the uncovered free space using IRIS, while treating previously computed regions as obstacles. This process is repeated until the desired coverage is met. We call this approach Iterative Obstacle Seeding ($\marcsthingy$).

In the following, all experiments are implemented in \href{https://drake.mit.edu/}{Drake}~\cite{drake}, and all computations are performed on a single desktop computer with an Intel Core i9-10850K CPU and 32 Gb of RAM. We solve all \maxclique instances to global optimality using Gurobi~\cite{gurobi}.

We evaluate VCC and IOS on four environments: \texttt{Village}, \texttt{3DOF Flipper}, \texttt{5DOF UR3}, and \texttt{7DOF IIWA}.
The dimension $n$ of these environments ranges from $3$ to $7$.
For each environment, we run the two algorithms ten times and report their performance in Table~\ref{tab:results}. 
The $\texttt{Village}$ environment from~\cite{marcucci2023fast} contains only convex obstacles and is compatible with the original IRIS~\cite{deits2015computing}. All other examples involve the configuration spaces of robotic manipulators and therefore IRIS-NP~\cite{petersen2023growing} is employed.
The number of samples used in the convergence check in Algorithm~\ref{alg:cseeding} is $M=5000$.

VCC meets the required coverage threshold with significantly fewer regions and in a substantially shorter amount of time. Notably, in the most challenging benchmark, \texttt{7DOF IIWA}, VCC requires 10 times fewer regions and meets the required coverage of $70\%$ around 16 times faster.
This substantial speedup can be attributed to two key factors.
First, the region inflation in VCC is parallelized.
Second, IRIS is the most computationally expensive step. While VCC only requires a single iteration of IRIS per region, $\marcsthingy$ can requires around five IRIS iterations per region, due to the initialization with a potentially uninformative spherical metric.


\section{Limitations of Approximating Convex Sets with Cliques}\label{sec:discussion}

Despite the strong performance of our algorithm, the hardness of solving {\sc{\textbf{Min $\alpha$-ApproxConvexCover}}} makes it possible to construct simple examples that highlight pitfalls of our heuristic approach.
In this section, we discuss holes in $\CFree$ which leads to one such pitfall that is particularly insightful.

\begin{figure}[t!]
\vspace{-0.2cm}
    \centering
    \begin{tikzpicture}[scale = 0.65,  every node/.style={transform shape}]
\begin{scope}[scale = 0.7]
\draw[very thick] (-3,0) -- (3,0)--(0,5.196)--cycle;
\draw[very thick] (-0.6,1.4) -- (0.6,1.4)--(0,2.44)--cycle;
\draw (0.8,3.8) -- (-1.4,0);
\draw[fill]  (0,1.7) ellipse (0.05 and 0.05);
\draw (-2.2,1.4) -- (2.2,1.4);
\draw (-0.8,3.8) -- (1.4,0);

\draw[fill=red, draw=none, opacity = 0.5] (-3,0) -- (-2.2,1.4) -- (-0.6,1.4) -- (-1.4,0);
\draw[fill=red, draw=none, opacity = 0.5] (-0.8,3.8) -- (0,5.2) -- (0.8,3.8) -- (0,2.4);
\draw[fill=red, draw=none, opacity = 0.5] (0.6,1.4) -- (2.2,1.4) -- (3,0) -- (1.4,0);
\draw[fill=black, opacity = 0.3] (-0.6,1.4) -- (0,2.4) -- (0.6,1.4) -- cycle;
\end{scope}

\begin{scope}[scale=0.7, shift={(-7,0)}]
\draw[very thick] (-3,0) -- (3,0)--(0,5.196)--cycle;
\node at (0.72,2.25) {\LARGE$\varepsilon$};
\draw[very thick] (-0.6,1.4) -- (0.6,1.4)--(0,2.44)--cycle;

\draw[fill]  (0,1.73) ellipse (0.05 and 0.05);
\draw[|<->|] (0.1708,2.6257) -- (0.8654,1.4704);
\draw (-2.2,1.4) -- (2.2,1.4);
\draw[|<->|] (3.2675,0.1067) -- (0.2,5.4);
\draw[fill=green, draw=none, opacity=0.3] (-3,0) -- (-2.2,1.4) -- (2.2,1.4) --(3,0)-- (-3,0);
\draw[fill=green, draw=none, opacity=0.1] (-3,0) -- (-2,1.7) -- (2,1.7) --(3,0)-- (-3,0);

\draw[fill=black, opacity = 0.3] (-0.6,1.4) -- (0,2.4) -- (0.6,1.4) -- cycle;
\draw[opacity = 0.2] (-2,1.7) -- (2,1.7);

\end{scope}



\node at (-3.2067,2.061) {1};



\draw[fill=red, draw=none, opacity = 0.2] (-2.1,0) -- (-1.4,1.2) -- (0,1.2) -- (-0.7,0);
\draw[fill=red, draw=none, opacity = 0.2] (-0.7,2.5) -- (0,3.6) -- (0.7,2.4) -- (0,1.2);
\draw[fill=red, draw=none, opacity = 0.2] (0,1.2) -- (1.4,1.2) -- (2.1,0) -- (0.7,0);
\draw[opacity = 0.2] (-1.4,1.2) -- (1.4,1.2);
\draw[opacity = 0.2] (-0.7,0) -- (0.7,2.4);
\draw[opacity = 0.2] (0.7,0) -- (-0.7,2.5);
\end{tikzpicture}
    \caption{Maximum cliques of infinitely dense visibility graphs can enclose holes, and do not necessarily correspond to collision-free convex sets. The largest collision-free convex region (green trapezoid) has a smaller area than the union of red parallelograms when $0<\varepsilon\leq 1 - \sqrt{5/6}$. In this case, the convex hull of the maximum clique must enclose the hole. 
    }
    \label{fig:Maxclique}
    \vspace{-0.5cm}
\end{figure}
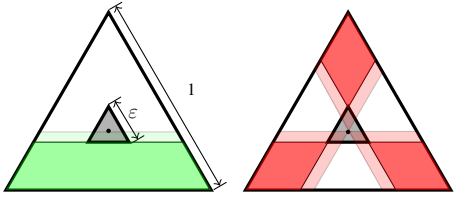

While every convex set in $\CFree$ naturally corresponds to a clique in the visibility graph, a clique in the visibility graph does not necessarily correspond to a convex set in $\CFree$. The convex hull of a clique can enclose holes.
This problem persists even if we sample arbitrarily dense visibility graphs, and if we restrict the analysis to maximum cliques.
A visual proof is shown in Figure~\ref{fig:Maxclique}, which illustrates a triangular configuration space with a triangular hole of size $\varepsilon$.
As $\varepsilon$ goes to zero, the largest convex subset of $\CFree$ is the green trapezoid, whose area approaches $5/9$ of the total area of $\CFree$.
On the other hand, a larger subset of mutually visible points is given by the union of the three red parallelograms, whose area approaches $6/9$ of the total area.
Therefore, if configurations are sampled uniformly at random, an optimum solution of \maxclique will almost surely enclose a hole as the number of samples goes to infinity.
A similar construction can be used to show an analogous discrepancy between the minimum convex cover of $\CFree$ and \mcc.

In principle, this problem can be addressed by solving a modified version of \maxclique that better captures the notion of a convex set.
In short, we require that every vertex of the visibility graph that is contained in the convex hull of the maximum clique must be a member of the clique.
For an infinitely dense visibility graph, this ensures that the maximum clique cannot enclose holes.

We enforce the contrapositive of the latter condition through linear constraints that separate all non-clique members $q_i$, with $i \in \{1, \ldots, K\}$, from the points in the clique with a hyperplane $ \calH_i = \{q \mid c_i^Tq +d_i = 0\}$, parameterized by the decision variables $(c_i, d_i)\inR^{n+1}$.  

The resulting mixed-integer linear optimization extends~\eqref{eqn:maxclique} by adding the additional separation constraints\vspace{-0.3cm}
\begin{subequations}
\label{eqn:maxcliqueconvhull}
\begin{align}
c_i^Tq_i + d_i &\geq 1-b_i, \quad &&\forall ~i = 1,\dots,K, \\
c_i^T q_j + d_i &\leq L (1 - b_j),
\quad &&\forall i, j = 1,\dots,K,
\end{align}
\end{subequations}
where $L$ is a large enough constant (e.g. a constant factor times the diameter of the visibility graph).  

\begin{figure}[t!]
\vspace{-0.2cm}
\centering
\includegraphics[width = 0.18\textwidth, trim = {0.1cm, 0.45cm, 0.1cm, 0.11cm}, clip]{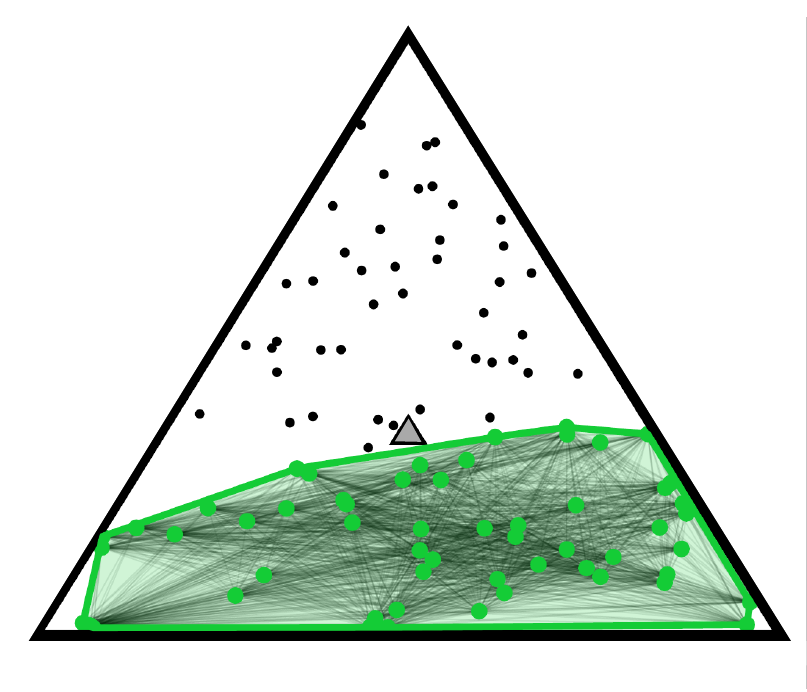}~~
\includegraphics[width = 0.18\textwidth, trim = {0.1cm, 0.45cm, 0.1cm, 0.11cm}, clip]{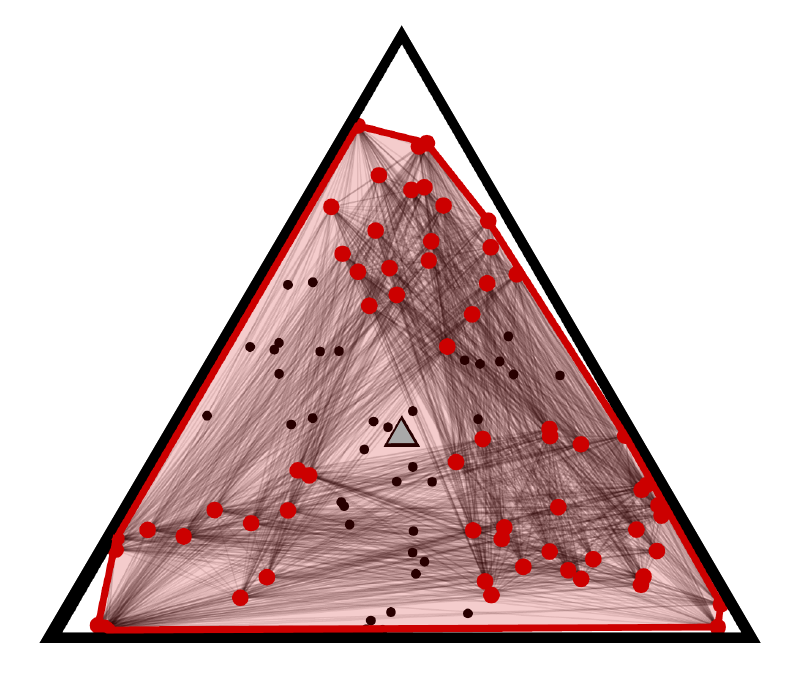}
\caption{A visibility graph with 100 random vertices in the triangular environment from Figure~\ref{fig:Maxclique}.
Solving the maximum clique problem~\eqref{eqn:maxclique} with the additional constraints~\eqref{eqn:maxcliqueconvhull} yields a clique with 56 vertices (shown in green), that closely approximates the corresponding convex set in Figure~\ref{fig:Maxclique}.
Solving only problem~\eqref{eqn:maxclique}, yields a clique with 63 vertices (shown in red) that, however, encloses the central hole.
\vspace{-0.5cm}}
    \label{fig:maxclique discrete}
\end{figure}

When the point $q_i$ is not in the clique ($b_i = 0$) and the point $q_j$ is in the clique ($b_j = 1$), these constraints read $c_i^Tq_i + d_i \geq 1$ and $c_i^T q_j + d_i \leq 0$.
Therefore, the hyperplane $\calH_i$ separates $q_i$ from $q_j$. On the other hand, these constraints are seen to be redundant for all other possible values of the binaries $b_{i}$ and $b_{j}$. In Figure \ref{fig:maxclique discrete}, we demonstrate how this extension prevents a maximum clique from enclosing holes in a finite-sample regime.

In practice, solving~\eqref{eqn:maxclique} with constraints~\eqref{eqn:maxcliqueconvhull} is too expensive for the problems in Table \ref{tab:results}.
Nonetheless, we observed that in the first \maxclique problem~\eqref{eqn:maxclique} only an average of $0.1\%$ of the vertices were excluded from the clique that could not be separated from it with a hyperplane.
Subsequent cliques, and improvements to the formulation~\eqref{eqn:maxcliqueconvhull}, demand a more nuanced discussion and are subject to future work. 
\section{Conclusions}

We have proposed an algorithm for approximately decomposing complex configuration spaces into small collections of polytopes.
Our algorithm uses clique covers of visibility graphs as an effective heuristic for obtaining local information about $\CFree$, and for seeding a region-inflation algorithm.
The parallels between convex sets in $\CFree$ and cliques in visibility graphs have also been discussed.
Our experiments demonstrate that VCC reliably finds approximate convex covers of $\CFree$ with fewer regions and in less time than previous approaches. 

\section*{Acknowledgements}
The authors would like to thank Steven M. LaValle, Thomas Cohn, Annan Zhang, and Fabian Dickhardt for the many fruitful discussions.

\IEEEtriggeratref{18}

\bibliographystyle{IEEEtran}
\bibliography{biblio.bib}

\begin{thebibliography}{10}
\providecommand{\url}[1]{#1}
\csname url@samestyle\endcsname
\providecommand{\newblock}{\relax}
\providecommand{\bibinfo}[2]{#2}
\providecommand{\BIBentrySTDinterwordspacing}{\spaceskip=0pt\relax}
\providecommand{\BIBentryALTinterwordstretchfactor}{4}
\providecommand{\BIBentryALTinterwordspacing}{\spaceskip=\fontdimen2\font plus
\BIBentryALTinterwordstretchfactor\fontdimen3\font minus
  \fontdimen4\font\relax}
\providecommand{\BIBforeignlanguage}[2]{{%
\expandafter\ifx\csname l@#1\endcsname\relax
\typeout{** WARNING: IEEEtran.bst: No hyphenation pattern has been}%
\typeout{** loaded for the language `#1'. Using the pattern for}%
\typeout{** the default language instead.}%
\else
\language=\csname l@#1\endcsname
\fi
#2}}
\providecommand{\BIBdecl}{\relax}
\BIBdecl

\bibitem{bishop2006pattern}
C.~M. Bishop and N.~M. Nasrabadi, \emph{Pattern recognition and machine
  learning}.\hskip 1em plus 0.5em minus 0.4em\relax Springer, 2006, vol.~4,
  no.~4.

\bibitem{pharr1997rendering}
M.~Pharr, C.~Kolb, R.~Gershbein, and P.~Hanrahan, ``Rendering complex scenes
  with memory-coherent ray tracing,'' in \emph{Proceedings of the 24th annual
  conference on Computer graphics and interactive techniques}, 1997, pp.
  101--108.

\bibitem{gilbert1988fast}
E.~G. Gilbert, D.~W. Johnson, and S.~S. Keerthi, ``A fast procedure for
  computing the distance between complex objects in three-dimensional space,''
  \emph{IEEE Journal on Robotics and Automation}, vol.~4, no.~2, pp. 193--203,
  1988.

\bibitem{marcucci2021shortest}
T.~Marcucci, J.~Umenberger, P.~A. Parrilo, and R.~Tedrake, ``Shortest paths in
  graphs of convex sets,'' \emph{arXiv preprint arXiv:2101.11565}, 2021.

\bibitem{marcucci2022motion}
T.~Marcucci, M.~Petersen, D.~von Wrangel, and R.~Tedrake, ``Motion planning
  around obstacles with convex optimization,'' \emph{arXiv preprint
  arXiv:2205.04422}, 2022.

\bibitem{cohn2023non}
T.~Cohn, M.~Petersen, M.~Simchowitz, and R.~Tedrake, ``Non-euclidean motion
  planning with graphs of geodesically-convex sets,'' \emph{arXiv preprint
  arXiv:2305.06341}, 2023.

\bibitem{kurtz2023temporal}
V.~Kurtz and H.~Lin, ``Temporal logic motion planning with convex optimization
  via graphs of convex sets,'' \emph{arXiv preprint arXiv:2301.07773}, 2023.

\bibitem{marcucci2023fast}
T.~Marcucci, P.~Nobel, R.~Tedrake, and S.~Boyd, ``Fast path planning through
  large collections of safe boxes,'' \emph{arXiv preprint arXiv:2305.01072},
  2023.

\bibitem{latombe2012robot}
J.-C. Latombe, \emph{Robot motion planning}.\hskip 1em plus 0.5em minus
  0.4em\relax Springer Science \& Business Media, 2012, vol. 124.

\bibitem{deits2015computing}
R.~Deits and R.~Tedrake, ``Computing large convex regions of obstacle-free
  space through semidefinite programming,'' in \emph{Algorithmic Foundations of
  Robotics XI}.\hskip 1em plus 0.5em minus 0.4em\relax Springer, 2015, pp.
  109--124.

\bibitem{amice2022finding}
A.~Amice, H.~Dai, P.~Werner, A.~Zhang, and R.~Tedrake, ``Finding and optimizing
  certified, collision-free regions in configuration space for robot
  manipulators,'' in \emph{Algorithmic Foundations of Robotics XV: Proceedings
  of the Fifteenth Workshop on the Algorithmic Foundations of Robotics}.\hskip
  1em plus 0.5em minus 0.4em\relax Springer, 2022, pp. 328--348.

\bibitem{petersen2023growing}
M.~Petersen and R.~Tedrake, ``Growing convex collision-free regions in
  configuration space using nonlinear programming,'' \emph{arXiv preprint
  arXiv:2303.14737}, 2023.

\bibitem{dai2023certified}
H.~Dai, A.~Amice, P.~Werner, A.~Zhang, and R.~Tedrake, ``Certified polyhedral
  decompositions of collision-free configuration space,'' \emph{arXiv preprint
  arXiv:2302.12219}, 2023.

\bibitem{simeon2000visibility}
T.~Sim{\'e}on, J.-P. Laumond, and C.~Nissoux, ``Visibility-based probabilistic
  roadmaps for motion planning,'' \emph{Advanced Robotics}, vol.~14, no.~6, pp.
  477--493, 2000.

\bibitem{lozano1979algorithm}
T.~Lozano-P{\'e}rez and M.~A. Wesley, ``An algorithm for planning
  collision-free paths among polyhedral obstacles,'' \emph{Communications of
  the ACM}, vol.~22, no.~10, pp. 560--570, 1979.

\bibitem{abrahamsen2022covering}
M.~Abrahamsen, ``Covering polygons is even harder,'' in \emph{2021 IEEE 62nd
  Annual Symposium on Foundations of Computer Science (FOCS)}.\hskip 1em plus
  0.5em minus 0.4em\relax IEEE, 2022, pp. 375--386.

\bibitem{eidenbenz2003approximation}
S.~J. Eidenbenz and P.~Widmayer, ``An approximation algorithm for minimum
  convex cover with logarithmic performance guarantee,'' \emph{SIAM Journal on
  Computing}, vol.~32, no.~3, 2003.

\bibitem{abrahamsen2023constructing}
M.~Abrahamsen, W.~Bille~Meyling, and A.~Nusser, ``Constructing concise convex
  covers via clique covers (cg challenge),'' in \emph{39th International
  Symposium on Computational Geometry (SoCG 2023)}.\hskip 1em plus 0.5em minus
  0.4em\relax Schloss Dagstuhl-Leibniz-Zentrum f{\"u}r Informatik, 2023.

\bibitem{da2023shadoks}
G.~D. da~Fonseca, ``Shadoks approach to convex covering,'' \emph{arXiv preprint
  arXiv:2303.07696}, 2023.

\bibitem{mamou2009simple}
K.~Mamou and F.~Ghorbel, ``A simple and efficient approach for 3d mesh
  approximate convex decomposition,'' in \emph{2009 16th IEEE international
  conference on image processing (ICIP)}.\hskip 1em plus 0.5em minus
  0.4em\relax IEEE, 2009, pp. 3501--3504.

\bibitem{varava2021free}
A.~Varava, J.~F. Carvalho, D.~Kragic, and F.~T. Pokorny, ``Free space of rigid
  objects: Caging, path non-existence, and narrow passage detection,''
  \emph{The international journal of robotics research}, vol.~40, no. 10-11,
  pp. 1049--1067, 2021.

\bibitem{kavraki1995computation}
L.~E. Kavraki, ``Computation of configuration-space obstacles using the fast
  fourier transform,'' \emph{IEEE Transactions on Robotics and Automation},
  vol.~11, no.~3, pp. 408--413, 1995.

\bibitem{branicky1990computing}
M.~Branicky and W.~Newman, ``Rapid computation of configuration space
  obstacles,'' in \emph{Proceedings., IEEE International Conference on Robotics
  and Automation}, 1990, pp. 304--310 vol.1.

\bibitem{sarmientoy2005sample}
A.~Sarmientoy, R.~Murrieta-Cidz, and S.~Hutchinsony, ``A sample-based convex
  cover for rapidly finding an object in a 3-d environment,'' in
  \emph{Proceedings of the 2005 IEEE International Conference on Robotics and
  Automation}.\hskip 1em plus 0.5em minus 0.4em\relax IEEE, 2005, pp.
  3486--3491.

\bibitem{karp2010reducibility}
R.~M. Karp, \emph{Reducibility among combinatorial problems}.\hskip 1em plus
  0.5em minus 0.4em\relax Springer, 2010.

\bibitem{strash2022effective}
D.~Strash and L.~Thompson, ``Effective data reduction for the vertex clique
  cover problem,'' in \emph{2022 Proceedings of the Symposium on Algorithm
  Engineering and Experiments (ALENEX)}.\hskip 1em plus 0.5em minus 0.4em\relax
  SIAM, 2022, pp. 41--53.

\bibitem{schwarzer2004exact}
F.~Schwarzer, M.~Saha, and J.-C. Latombe, ``Exact collision checking of robot
  paths,'' \emph{Algorithmic foundations of robotics V}, pp. 25--41, 2004.

\bibitem{lavalle2006planning}
S.~M. LaValle, \emph{Planning algorithms}.\hskip 1em plus 0.5em minus
  0.4em\relax Cambridge university press, 2006.

\bibitem{amice2023certifying}
A.~Amice, P.~Werner, and R.~Tedrake, ``Certifying bimanual rrt motion plans in
  a second,'' \emph{arXiv preprint arXiv:2310.16603}, 2023.

\bibitem{boyd2004convex}
S.~P. Boyd and L.~Vandenberghe, \emph{Convex optimization}.\hskip 1em plus
  0.5em minus 0.4em\relax Cambridge university press, 2004.

\bibitem{kuffner2000rrt}
J.~J. Kuffner and S.~M. LaValle, ``{RRT}-connect: An efficient approach to
  single-query path planning,'' in \emph{Proceedings 2000 ICRA. Millennium
  Conference. IEEE International Conference on Robotics and Automation.
  Symposia Proceedings (Cat. No. 00CH37065)}, vol.~2.\hskip 1em plus 0.5em
  minus 0.4em\relax IEEE, 2000, pp. 995--1001.

\bibitem{kavraki1996probabilistic}
L.~E. Kavraki, P.~Svestka, J.-C. Latombe, and M.~H. Overmars, ``Probabilistic
  roadmaps for path planning in high-dimensional configuration spaces,''
  \emph{IEEE transactions on Robotics and Automation}, vol.~12, no.~4, pp.
  566--580, 1996.

\bibitem{drake}
\BIBentryALTinterwordspacing
R.~Tedrake and the Drake Development~Team, ``Drake: Model-based design and
  verification for robotics,'' 2019. [Online]. Available:
  \url{https://drake.mit.edu}
\BIBentrySTDinterwordspacing

\bibitem{gurobi}
{Gurobi Optimization, LLC}, ``\href{https://www.gurobi.com/}{Gurobi Optimizer
  Reference Manual},'' 2023.

\end{thebibliography}


\end{document}